\documentclass[sigconf]{acmart}
\usepackage{multirow}

\renewcommand\footnotetextcopyrightpermission[1]{} 
\AtBeginDocument{%
  \providecommand\BibTeX{{%
    \normalfont B\kern-0.5em{\scshape i\kern-0.25em b}\kern-0.8em\TeX}}}

\setcopyright{acmlicensed}
\copyrightyear{2018}
\acmYear{2018}
\acmDOI{XXXXXXX.XXXXXXX}

\acmConference[Conference acronym 'XX]{Make sure to enter the correct
  conference title from your rights confirmation emai}{June 03--05,
  2018}{Woodstock, NY}
\acmISBN{978-1-4503-XXXX-X/18/06}

\begin{document}

\title{Diffutoon: High-Resolution Editable Toon Shading via Diffusion Models}

\author{Zhongjie Duan}
\affiliation{
  \institution{East China Normal University}
  \city{Shanghai}
  \country{China}}
\email{zjduan@stu.ecnu.edu.cn}

\author{Chengyu Wang}
\affiliation{
  \institution{Alibaba Group}
  \city{Hangzhou}
  \country{China}}
\email{chengyu.wcy@alibaba-inc.com}

\author{Cen Chen}
\affiliation{
  \institution{East China Normal University}
  \city{Shanghai}
  \country{China}}
\email{cenchen@dase.ecnu.edu.cn}

\author{Weining Qian}
\affiliation{
  \institution{East China Normal University}
  \city{Shanghai}
  \country{China}}
\email{wnqian@dase.ecnu.edu.cn}

\author{Jun Huang}
\affiliation{
  \institution{Alibaba Group}
  \city{Hangzhou}
  \country{China}}
\email{huangjun.hj@alibaba-inc.com}

\renewcommand{\shortauthors}{Duan et al.}

\begin{abstract}
  Toon shading is a type of non-photorealistic rendering task of animation. Its primary purpose is to render objects with a flat and stylized appearance. As diffusion models have ascended to the forefront of image synthesis methodologies, this paper delves into an innovative form of toon shading based on diffusion models, aiming to directly render photorealistic videos into anime styles. In video stylization, extant methods encounter persistent challenges, notably in maintaining consistency and achieving high visual quality. In this paper, we model the toon shading problem as four subproblems: stylization, consistency enhancement, structure guidance, and colorization. To address the challenges in video stylization, we propose an effective toon shading approach called \textit{Diffutoon}. Diffutoon is capable of rendering remarkably detailed, high-resolution, and extended-duration videos in anime style. It can also edit the content according to prompts via an additional branch. The efficacy of Diffutoon is evaluated through quantitive metrics and human evaluation. Notably, Diffutoon surpasses both open-source and closed-source baseline approaches in our experiments. Our work is accompanied by the release of both the source code and example videos on Github\footnote{Project page: \url{https://ecnu-cilab.github.io/DiffutoonProjectPage/}}.
\end{abstract}





\maketitle

\section{Introduction}

Toon shading \cite{barla2006x} is a crucial task within the animation industry, aiming to render 3D computer-generated graphics in a flat style. These techniques are extensively applied across diverse domains, including video game development and animation production \cite{hudon20182d}. As diffusion models \cite{sohl2015deep} achieve impressive performance in image synthesis, we discern their potential in video stylization. In this paper, we explore a new type of toon shading task, aiming to directly transform photorealistic videos into an animated visual style.

In recent years, Stable Diffusion \cite{rombach2022high}, a diffusion model pre-trained on large-scale text-image datasets \cite{schuhmann2022laion}, has emerged as a powerful backbone in text-to-image synthesis. In open-source communities, abundant fine-tuned models based on Stable Diffusion are available to handle diverse styles. Nevertheless, extending diffusion models to video processing presents many challenges \cite{xing2023survey}. First, there is a lack of controllability. When applying diffusion models to videos, it is difficult to retain essential information in the original video, such as structure and lighting. Second, the consistency issue is crucial, as independently processing each frame often leads to undesirable flickering. Third, visual quality remains a concern. While video platforms commonly support resolutions up to 1080P and even 4K, most diffusion models struggle to process high-resolution videos.

Prior studies have attempted to address these challenges. In controllable image synthesis, adapter-type control modules \cite{zhang2023adding,mou2023t2i} have already demonstrated the capability for precise control. However, these modules are limited to processing individual images and cannot handle videos. To improve video consistency, studies on this topic are typically categorized into two types: training-free and training-based approaches. Training-free methods \cite{yang2023rerender,ceylan2023pix2video} align content between frames by constructing specific mechanisms, requiring no training process, but their effectiveness is limited. On the other hand, training-based methods \cite{esser2023structure,guo2023animatediff} can generally achieve better results. However, due to the substantial computational resources required, training diffusion models on lengthy video datasets remains exceedingly challenging. Consequently, most video diffusion models can only handle up to a maximum of 32 continuous frames, leading to inconsistencies in longer videos. To achieve better visual quality, super-resolution techniques \cite{wang2021real} can potentially enhance video resolution, but they may introduce extra issues like over-smoothed information loss \cite{li2022srdiff}.

In this paper, we propose a video processing method specifically designed for toon shading. We divide the toon shading problem into four subproblems: stylization, consistency enhancement, structure guidance, and colorization. For each subproblem, we provide a specific solution. Our proposed framework consists of a main toon shading pipeline and an editing branch. In the main toon shading pipeline, we construct a multi-module denoising model based on an anime-style diffusion model. ControlNet \cite{zhang2023adding} and AnimateDiff \cite{guo2023animatediff} are utilized in the denoising model to address controllability and consistency issues. To enable the generation of ultra-high-resolution content in long videos, we depart from the conventional frame-by-frame generation paradigm. Instead, we adopt a sliding window approach to iteratively update the latent embedding of each frame. Additionally, our method offers the capability to edit videos through the editing branch, which provides editing signals for the main toon shading pipeline. To improve the efficiency, we incorporate flash attention \cite{dao2022flashattention} into the attention mechanisms, effectively mitigating excessive GPU memory usage. Remarkably, our approach can directly handle resolutions of up to $1536\times 1536$. In our experiments, we first evaluate Diffutoon in the toon shading task, and then we evaluate the capability of editing some content according to given prompts. Comparative analyses are conducted with both open-source and closed-source approaches. Quantitative assessments and human evaluations consistently demonstrate the significant advantages of our approach over other methods. The contribution of this paper includes:
\begin{itemize}
    \item We introduce an innovative form of toon shading, aiming to release the potential of generative diffusion models in the field of non-photorealistic rendering.
    \item We propose an effective toon shading approach based on diffusion models, making it possible to transform photorealistic videos into an anime style and edit the content according to given prompts if required.
    \item Our implementation presents a robust framework for deploying diffusion models in video processing. This framework can achieve very high resolution and is capable of processing long videos.
\end{itemize}

\section{Related Work}


\subsection{Stable Diffusion}


Stable Diffusion \cite{rombach2022high} has emerged as a popular foundational backbone within both the academic and open-source communities. Its structure includes a text encoder \cite{radford2021learning}, a UNet \cite{ronneberger2015u}, and a VAE \cite{kingma2013auto}. To leverage Stable Diffusion models effectively for toon shading applications, a specialized anime-style image generation model tailored for image-to-image processing is essential. By employing advanced training methods such as LoRA \cite{hu2021lora}, Textual Inversion \cite{gal2022image}, DreamBooth \cite{ruiz2023dreambooth}, and others, we can easily fine-tune a personalized model. Additionally, the utilization of prompt engineering techniques \cite{cao2023beautifulprompt} allows for the refinement of prompts, thereby enabling the generation of high-aesthetic images.

\subsection{Fast Sampling of Diffusion Models}


Diffusion models typically require multiple iterative steps to generate clear images, making their generation speed comparatively slower than that of GANs \cite{goodfellow2014generative}. In video processing, where each frame needs to be processed, the issue of computational efficiency becomes even more significant. Some studies \cite{song2020denoising,lu2022dpm,duan2023olss} have addressed this by introducing schedulers to control the generation process, making it possible to generate clear images in a few steps. In high-resolution image generation, although some existing research \cite{jin2023training,he2023scalecrafter} has showcased the feasibility of transferring low-resolution models to high-resolution tasks, the computational load of attention layers in high-resolution image generation remains a concern. To alleviate this issue, efficient attention implementations such as flash attention \cite{dao2022flashattention} have reduced the memory and time requirements, enabling the processing of high-resolution videos.

\subsection{Controllable Image Synthesis}


To enhance the controllability of the generated results in diffusion models, recent studies such as ControlNet \cite{zhang2023adding} and T2I-Adapter \cite{mou2023t2i} aim to integrate control signals into the generation process. By connecting controlling modules in the form of adapters to the UNet, we can construct a robust image-to-image pipeline and selectively retain information from the original image. The advancements in controllable image-to-image techniques inspired the studies in video-to-video generation. For instance, Gen-1 \cite{esser2023structure} decomposes video information into structural and content components. It leverages depth estimation \cite{ranftl2020towards} to represent the structural details and synthesize a stylized video. In this paper, we reference and adopt similar controlling strategies in our proposed method.

\subsection{Temporal Diffusion Models}


The primary challenge in the application of diffusion models to video processing is consistency. The conventional practice of independently processing each frame invariably results in video flickering. Some studies \cite{khachatryan2023text2video,yang2023rerender} address this issue by incorporating special mechanisms, such as cross-frame attention, which aligns the content of adjacent frames without the need for training. Other studies \cite{blattmann2023align,esser2023structure,guo2023animatediff} tackle the consistency problem by introducing trainable modules and training them on video datasets. Among these studies, AnimateDiff \cite{guo2023animatediff}, being compatible with Stable Diffusion architecture, has gained significant popularity within open-source communities. In our methodology, we utilize motion modules based on AnimateDiff to enhance the overall coherence.

\subsection{Post-Processing Methods}


Training diffusion models on long videos still faces challenges due to the high computational resource requirements. Some video post-processing approaches can be employed to assist in enhancing the long-term consistency of videos. For instance, CoDeF \cite{ouyang2023codef}, FastBlend \cite{duan2023fastblend}, and other blind video deflickering algorithms \cite{lei2023blind}. While these methods can handle longer videos, they typically encounter issues such as screen tearing and blurring when dealing with high-speed and substantial motion. The method proposed in this paper draws inspiration from such approaches to improve long-term consistency.

\section{Methodology}

\begin{figure*}[htbp]
  \centering
  \includegraphics[width=.98\linewidth]{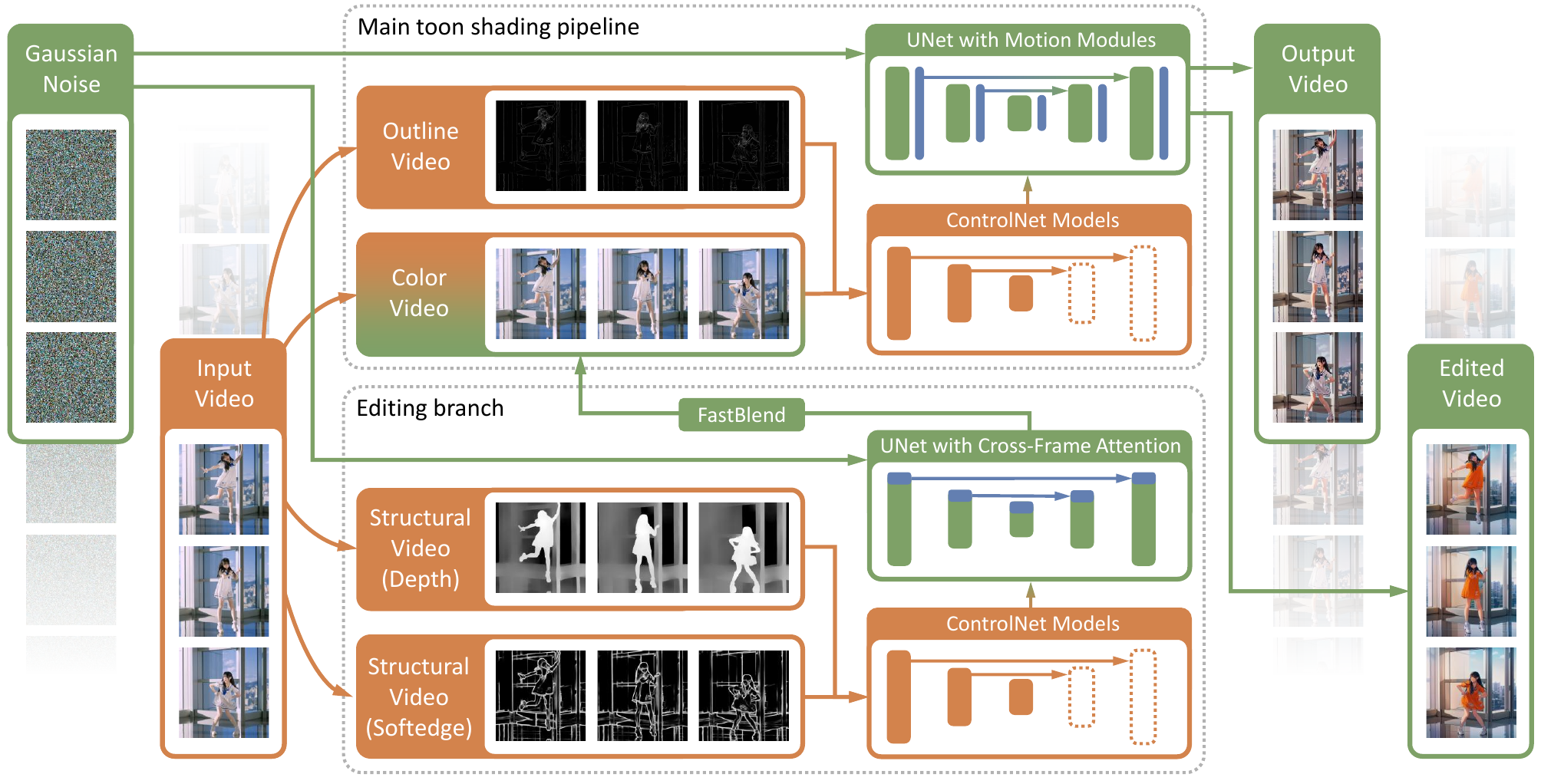}
  \caption{The overall architecture of Diffutoon, where the top part is the main toon shading pipeline, and the bottom part is the editing branch. The editing branch can generate editing signals in the format of color video for the main toon shading pipeline.}
  \label{figure:main_figure}
\end{figure*}

The overall architecture of Diffutoon is illustrated in Figure \ref{figure:main_figure}. The whole approach consists of a main toon shading pipeline and an editing branch. The main toon shading pipeline can render the input video in an anime style. To enable anime video editing, we designed an additional editing branch to generate an edited color video for the main toon shading pipeline.

\subsection{Toon Shading}

We divide the toon shading task into four subtasks: stylization, consistency enhancement, structure guidance, and colorization. We employ four models to address the four subtasks respectively.

\begin{itemize}
  \item \textbf{Stylization}: we leverage a personalized Stable Diffusion \cite{rombach2022high} model for anime stylization. Theoretically, our approach supports every open-sourced diffusion model with such model architecture.
  \item \textbf{Consistency enhancement}: to enhance the temporal consistency, we employ several motion modules in our approach. These modules are based on AnimateDiff \cite{guo2023animatediff}, which are inserted into the UNet to keep the content consistent.
  \item \textbf{Structure guidance}: we extract the outline information from the input video and use a ControlNet model to retain the outline information during the generation process. Unlike some existing methods \cite{esser2023structure} that use depth estimation to represent structural information, we employ outline as structural information, which is more suitable for rendering flat-style animations.
  \item \textbf{Colorization}: we use another ControlNet model for colorization. This model is trained for super-resolution tasks, thus it can improve the overall video quality even if the input video is in low resolution. This model directly processes the input video in the main toon shading pipeline, and it takes the edited color video as input when the editing branch is enabled.
\end{itemize}

As illustrated in the top part of Figure \ref{figure:main_figure}, the main toon shading pipeline involves several key steps. Given an input video containing $N$ frames $\{\boldsymbol v^1,\boldsymbol v^2,\cdots,\boldsymbol v^N\}$, we first generate a structural video and a color video. The structural video $\{\boldsymbol o^1,\boldsymbol o^2,\cdots,\boldsymbol o^N\}$ contains the outline information extracted from the input video, and the color video $\{\boldsymbol c^1,\boldsymbol c^2,\cdots,\boldsymbol c^N\}$ is the input video when the editing branch is disabled. Subsequently, the two videos serve as inputs to their respective ControlNet models, which in turn produce conditioning signals for the UNet. Simultaneously, the motion modules generate temporal signals. These four models constitute a large denoising model $\mathcal E$, employed iteratively to synthesize a visually consistent video.

In the denoising process, initially, the latent embedding of each frame is sampled from a Gaussian distribution.
\begin{equation}
    \boldsymbol x_T=\{\boldsymbol x^i_T\}_{i=1}^N \sim \mathcal N(\boldsymbol O,\boldsymbol I),
\end{equation}
where $T$ is the number of iterative steps and each embedding is independent identically distributed. At each denoising step, we use classifier-free guidance \cite{ho2021classifier} to build a textual guidance mechanism, which consists of a positive side and a negative side. On the positive side, we use some empirical keywords (e.g., ``best quality'', ``perfect anime illustration'') as prompt $\tau$ for better aesthetics. Note that the motion modules are trained within 32 consecutive frames, we can only use the denoising model $\mathcal E$ in a sliding window with a size no larger than 32. The sliding windows with size $d$ and stride $s$ are
\begin{equation}
    \mathcal W(d,s)=\left\{[i,i+d-1]:1\le i\le N,i\equiv1 (\text{mod }s)\right\},
\end{equation}
where $s<d$ for a smooth transition between different sliding windows. In a sliding window $[l,r]$, The model output on the positive side is
\begin{equation}
    \left\{\boldsymbol e_{t,+}(l,i,r)\right\}_{i=l}^r=\mathcal E\Big(
    \left\{\boldsymbol x_t^i\right\}_{i=l}^r,
    \left\{\boldsymbol o_t^i\right\}_{i=l}^r,
    \left\{\boldsymbol c_t^i\right\}_{i=l}^r,
    t,
    \tau
    \Big).
    \label{equation:model_output}
\end{equation}
The latent embeddings are initially stored in RAM and will be moved to GPU memory when the sliding window covers them. We adopt a linear combination of overlapping segments from different sliding windows.
\begin{equation}
    \overline{\boldsymbol e}_{t,+}(i)=\sum_{(l,r)\in\mathcal W(d,s)}\frac{w(l,i,r)}{\sum_{(l',r')\in\mathcal W(d,s)}w(l',i,r')} \boldsymbol e_{t,+}(l,i,r).
    \label{equation:sliding_window}
\end{equation}
The weight $w(l,i,r)$ is formulated as follows:
\begin{equation}
    w(l,i,r)=\begin{cases}
    1+\epsilon-\left|i-\frac{l+r}{2}\right|/\frac{r-l}{2}, &\text{if }l\le i\le r,\\
    0, &\text{otherwise},
    \end{cases}
\end{equation}
where $\epsilon=10^{-2}$ is the minimum weight of tailed frames. This allows the information from each frame to be shared with other frames throughout the generation process. This mechanism implicitly implements a large size of sliding window, enhancing the long-term consistency of generated content. To avoid disintegrated parts on faces and hands, we employ a textual inversion $\tau'$ \cite{gal2022image} on the negative side, which involves 10 token embeddings to be processed by the text encoder. By replacing $\tau$ with $\tau'$ in (\ref{equation:model_output}) and (\ref{equation:sliding_window}), we can obtain the estimated noise on the negative side $\overline{\boldsymbol e}_{t,-}(i)$. Then, the guided estimated noise is
\begin{equation}
    \overline{\boldsymbol e}_t(i)=g\cdot\overline{\boldsymbol e}_{t,+}(i)+(1-g)\cdot\overline{\boldsymbol e}_{t,-}(i).
\end{equation}
The classifier-free guidance scale $g$ is set to 7 by default. Based on empirical evidence, we skip the final attention layer of the text encoder, which can improve the visual quality slightly. The overall estimated noise of the whole video is
\begin{equation}
    \overline{\boldsymbol e}_t=\big(\overline{\boldsymbol e}(0),\overline{\boldsymbol e}(1),\cdots,\overline{\boldsymbol e}(n)\big)\in \mathbb R^{N\times H\times W\times C}.
\end{equation}
After that, we utilize a DDIM \cite{song2020denoising} scheduler to control the generation process.
\begin{equation}
    \boldsymbol x_{t-1}=\sqrt{\alpha_{t-1}}\left(
    \frac{\boldsymbol x_t-\sqrt{1-\alpha_t}\overline{\boldsymbol e}_t}{\sqrt{\alpha_t}}
    \right)
    +\sqrt{1-\alpha_{t-1}}\overline{\boldsymbol e}_t,
\end{equation}
where $\alpha_t$ is the hyper-parameter that determines how much noise it contains in step $t$. We follow the implementation of DDIM in AnimateDiff \cite{guo2023animatediff}. Despite the findings from recent studies suggesting that alternative schedulers, such as DPM-Solver \cite{lu2022dpm} and OLSS \cite{duan2023olss}, can achieve superior visual quality within a specified number of steps, we decide to employ such a straightforward scheduler due to memory constraints. This decision is driven by the fact that these alternative schedulers need to store all latent tensors throughout the generation process, posing challenges for processing long videos. Additionally, we set the number of denoising steps $T$ to only 10 for faster generation without compromising the resulting quality.

\subsection{Adding Editing Signals to Toon Shading}
\label{subsection:editing_branch}

In the main toon shading pipeline, we decompose the information in the input video into outlines and colors. In practice, we can edit the content by modifying the outline video or color video. Notably, due to the lack of reliable video editing methods for structural information, we mainly focus on editing the color information. We observe that the ControlNet model used for processing color videos can assist the UNet in generating high-quality videos, even if the color videos are blurry. This noteworthy insight implies a robust fault tolerance within our approach to video editing methods. Consequently, we are motivated to design a dedicated branch to support video editing.

To achieve this, we add an editing branch to generate text-guided editing signals for the main toon shading pipeline, where the editing signal is passed in the format of a color video. The architecture of the editing branch is shown in the bottom part of Figure \ref{figure:main_figure}. Similar to the main toon shading pipeline, we divide the synthesis of the editing signal into four subtasks:
\begin{itemize}
    \item \textbf{Stylization}: we leverage the same Stable Diffusion model as that in the main toon shading pipeline.
    \item \textbf{Consistency enhancement}: we use cross-frame attention and FastBlend \cite{duan2023fastblend} to improve consistency. While the motion modules based on AnimateDiff can make the video fluent, there are instances where they compromise visual quality. This pitfall is due to their reliance on a modified DDIM scheduler, which will be further discussed in the following experiments. This is also the reason why a single editing branch cannot synthesize a high-quality video. To release the potential of the diffusion model itself, we use the DDIM scheduler consistent with its training process, omitting these motion modules. Instead, we leverage cross-frame attention and FastBlend to improve consistency, where cross-frame attention is a widely demonstrated effective technique \cite{yang2023rerender,ceylan2023pix2video}, and FastBlend is a model-free deflickering approach for post-processing.
    \item \textbf{Structure guidance}: we employ depth estimation \cite{ranftl2020towards} and softedge \cite{xie2015holistically} to represent the structural information and use two ControlNet models for precise structure guidance. Previous studies \cite{esser2023structure,duan2023diffsynth} have empirically demonstrated the efficacy of these configurations in preserving structural information, particularly in instances of significant video editing.
    \item \textbf{Colorization}: the color is determined by the given prompt. Note that sometimes the classifier-free guidance mechanism fails to generate the correct color in several frames. In such instances, FastBlend serves as a corrective measure, leveraging information from neighboring frames to rectify deficient color.
\end{itemize}
The other components of the editing branch are similar to those of the main toon shading pipeline. The same sliding window mechanism is applied on this branch. While the color video synthesized by this branch may exhibit blurring, it maintains a high level of visual coherence, suitable for guiding the main toon shading pipeline to synthesize a high-quality video.

\subsection{Synthesizing High-Resolution Long Videos}

We implement Diffutoon based on the DiffSynth framework \cite{duan2023diffsynth}, which can process the whole video in the latent space. To reduce the required GPU memory and improve computational efficiency, we adopt flash attention \cite{dao2022flashattention} in all attention layers, including the text encoder, UNet, VAE, ControlNet models, and motion modules. This memory-efficient attention implementation empowers the direct synthesis of videos in exceptionally high resolution. The sliding window mechanism is capable of extending the length of videos. With the above settings, our pipeline succeeds in synthesizing remarkably detailed, high-resolution, and extended-duration videos.

\section{Experiments}

Our primary focus centers on the rendering of high-resolution videos with rapid and substantial motion. To evaluate the efficacy of our proposed approach, we create a dataset comprising 10 videos sourced from a video platform\footnote{\url{https://www.bilibili.com/}}. This dataset will be released publicly. In our experiments, we achieve a video resolution of up to $1536\times 1536$, resulting in visually impressive frames. The detailed settings of models and parameters are presented in the appendix.

\subsection{Comparison with Baseline Methods}

The evaluation involves two distinct tasks: toon shading, where we exclusively employ the main toon shading pipeline to transform input videos into an anime style, and toon shading with editing signals, where manually crafted editing prompts are used to edit the content during the rendering process. In the two tasks, we conduct comparative evaluations with other state-of-the-art methods, including Rerender-a-video \cite{yang2023rerender}, an open-source method that utilizes a special pipeline for video synthesis. To ensure a fair comparison, we replace the default model of Rerender-a-video with the diffusion model from our approach. Additionally, we involve several popular closed-source models that have demonstrated competitiveness in comparison to existing methods. These models include Gen-1 \cite{esser2023structure} and DomoAI \cite{domoai2023}. Gen-1, while not specifically tailored for toon shading, is evaluated exclusively in the second task. DomoAI offers several models for users on Discord\footnote{\url{https://discord.com/}}, and in our experiments, we employ the ``Anime v2 - Japanese anime style'' version. Due to the length limitation of DomoAI, we only use 10 seconds from each video in our experiments. This comprehensive comparative analysis aims to evaluate the performance of our approach relative to both open-source and closed-source state-of-the-art methods across diverse tasks.

\begin{table}[t]
\centering
\tabcolsep=0.25em
\begin{tabular}{l|l|ccc}
\hline
\multirow{2}{*}{Task} & \multirow{2}{*}{Method} & \multicolumn{3}{c}{Metric}                    \\ \cline{3-5} 
                      &                        
& \begin{tabular}[c]{@{}c@{}}Aesthetic\\ score $\uparrow$\end{tabular} 
& \begin{tabular}[c]{@{}c@{}}CLIP\\ score $\uparrow$\end{tabular} 
& \begin{tabular}[c]{@{}c@{}}Pixel\\ MSE $\downarrow$\end{tabular} 
\\ \hline
\multirow{3}{*}{Toon shading}
& Rerender-a-video        & 5.35            & -              & 200.46          \\
& DomoAI                  & 6.26            & -              & -               \\
& Diffutoon               & \textbf{6.47}   & -              & \textbf{188.87} \\ \hline
\multirow{4}{*}{\begin{tabular}[c]{@{}l@{}}Toon shading\\ with editing\\ signals\end{tabular}}
& Rerender-a-video        & 5.40            & 28.63          & 266.23          \\
& DomoAI                  & 6.25            & 29.01          & -               \\
& Gen-1                   & 6.11            & 28.91          & -               \\
& Diffutoon               & \textbf{6.37}   & \textbf{30.69} & \textbf{143.51} \\ \hline
\end{tabular}
\caption{Quantitative results of each approach.}
\label{table:quantitative_results}
\end{table}

\begin{figure*}[htbp]
  \centering
  \begin{tabular}{cc}
  \includegraphics[width=0.475\linewidth]{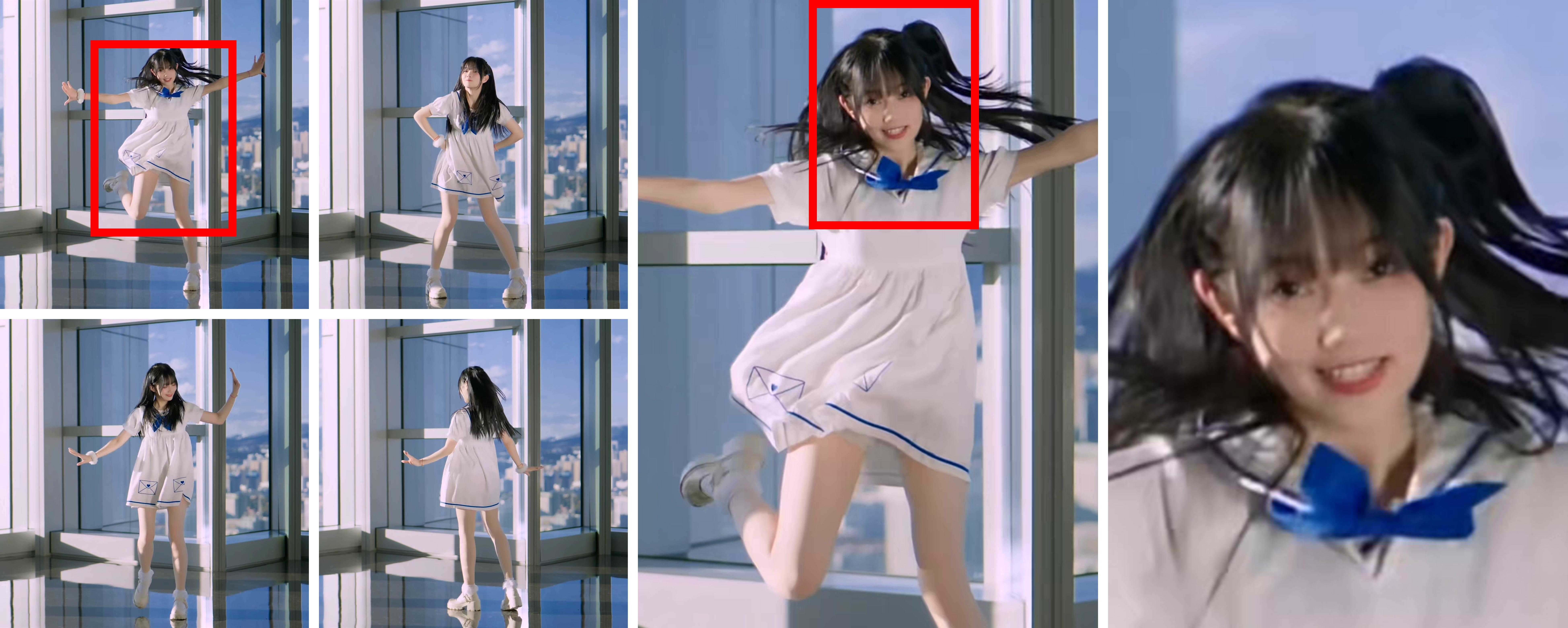} &
  \includegraphics[width=0.475\linewidth]{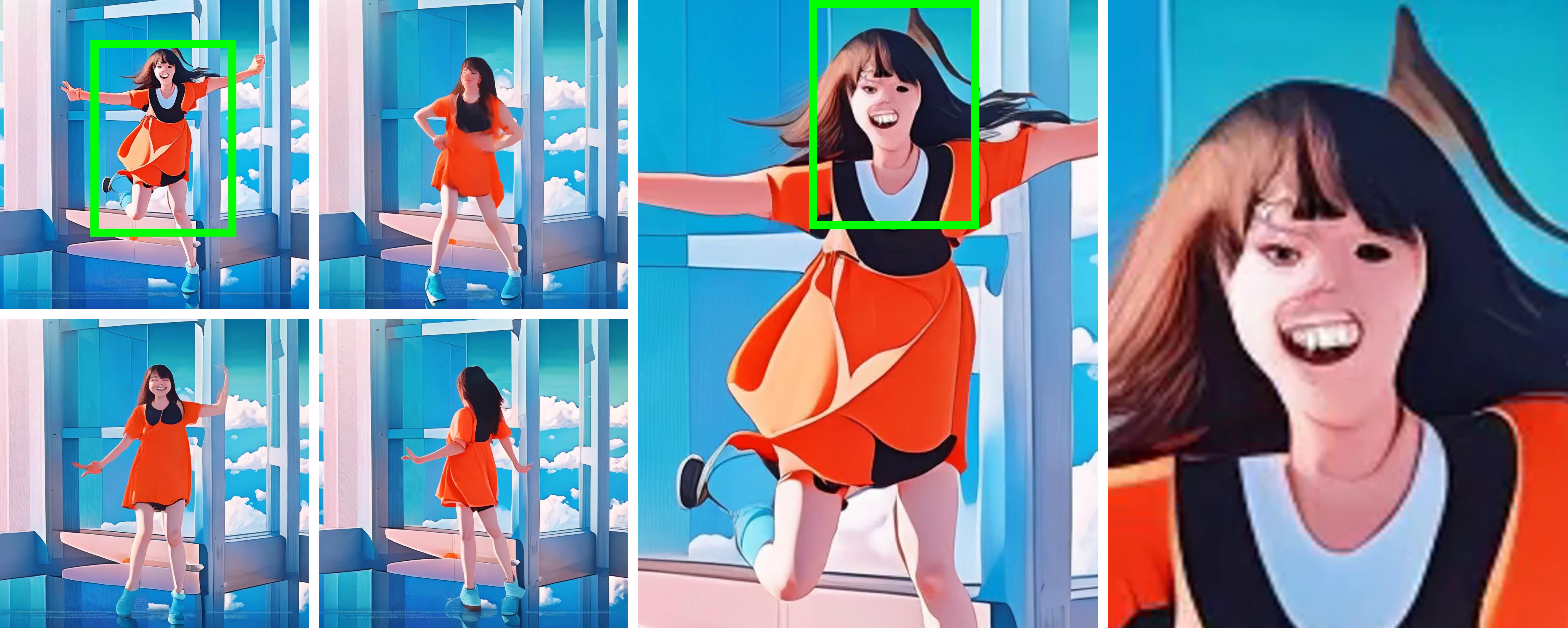} \\
  (a) Input video & (b) Gen-1 (toon shading with editing signals) \\
  \includegraphics[width=0.475\linewidth]{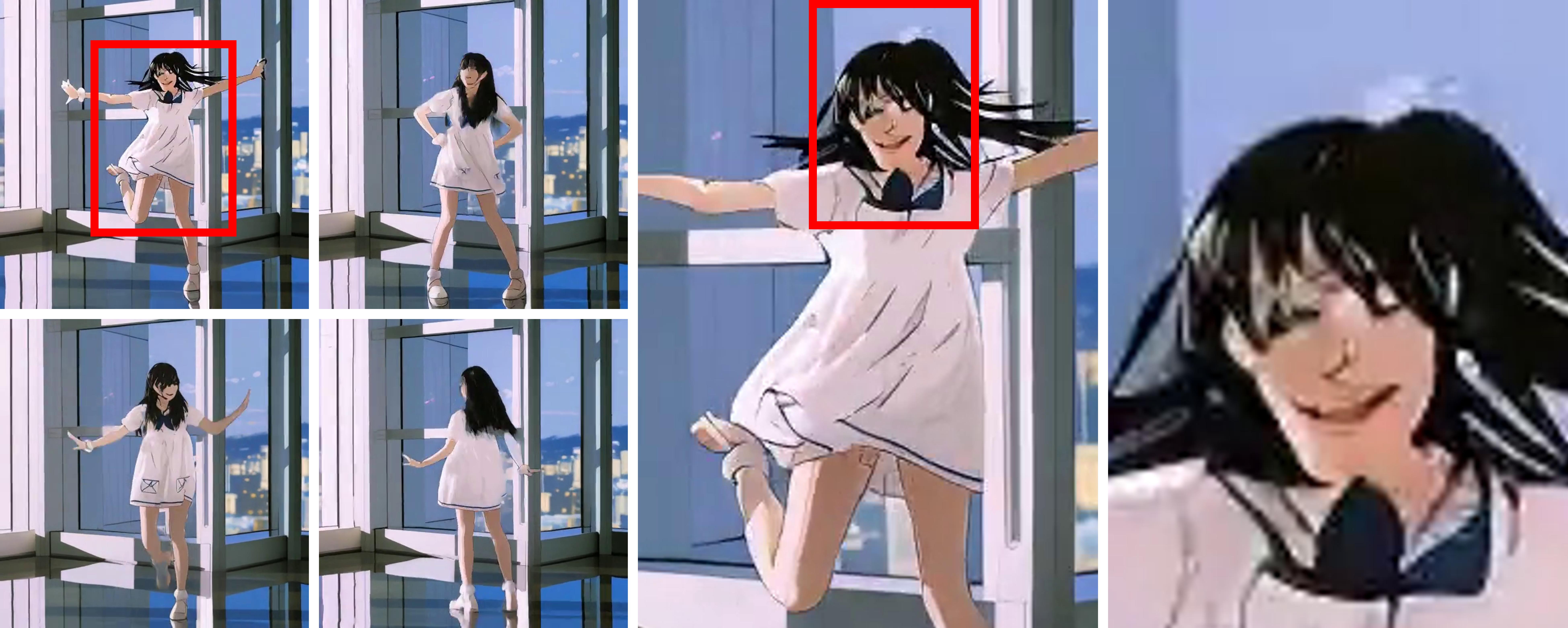} &
  \includegraphics[width=0.475\linewidth]{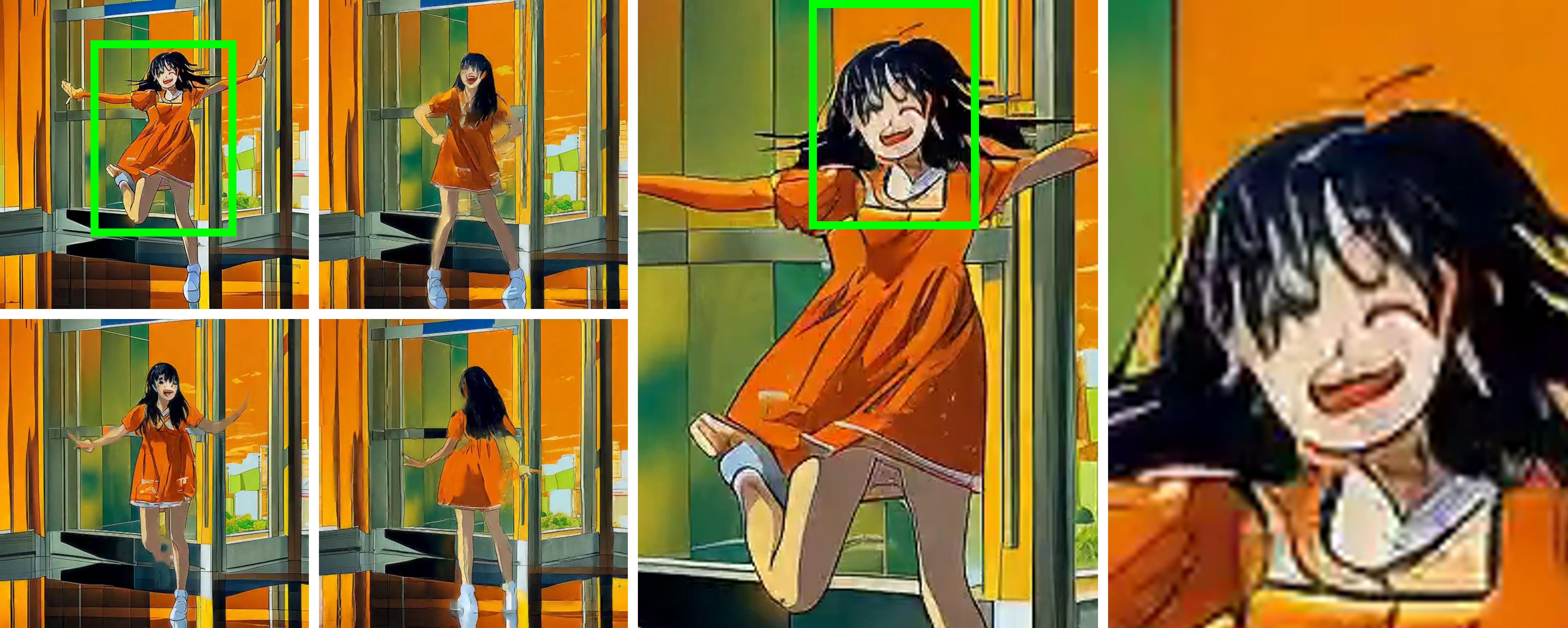} \\
  (c) Rerender-a-video (toon shading) & (d) Rerender-a-video (toon shading with editing signals) \\
  \includegraphics[width=0.475\linewidth]{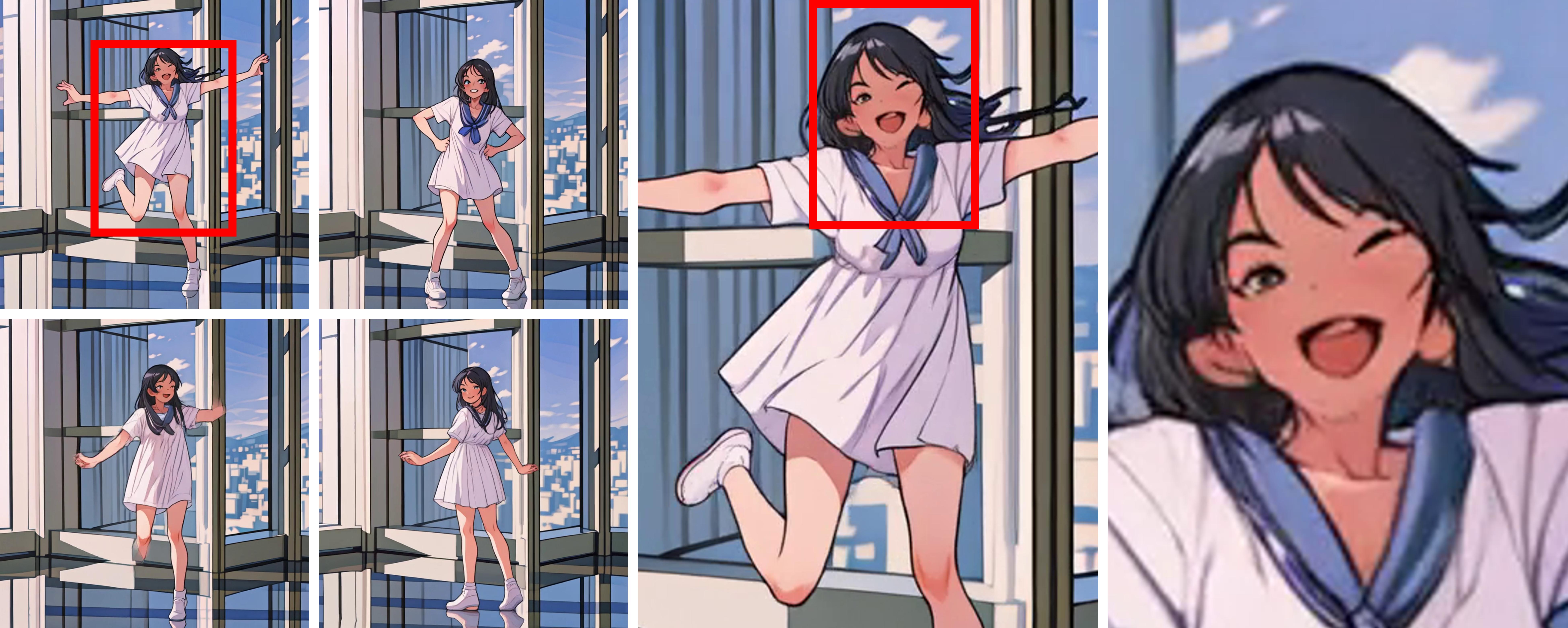} &
  \includegraphics[width=0.475\linewidth]{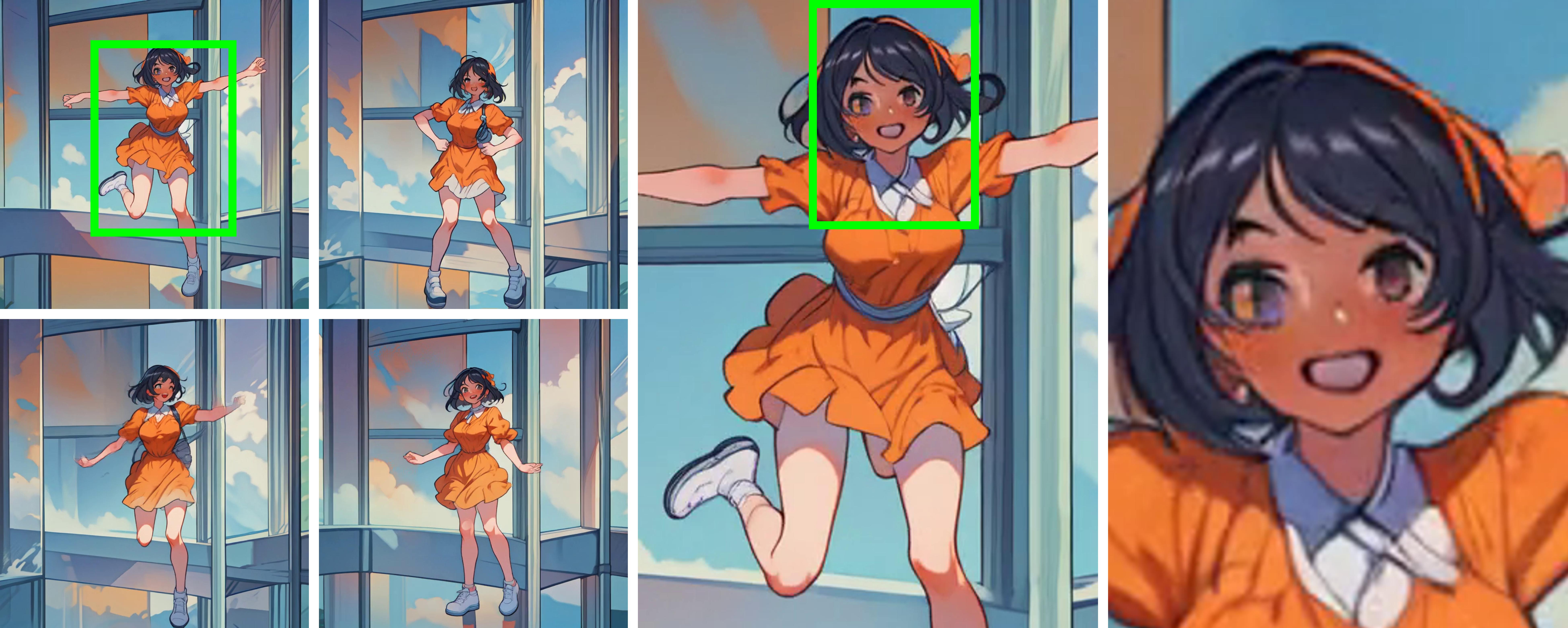} \\
  (e) DomoAI (toon shading) & (f) DomoAI (toon shading with editing signals) \\
  \includegraphics[width=0.475\linewidth]{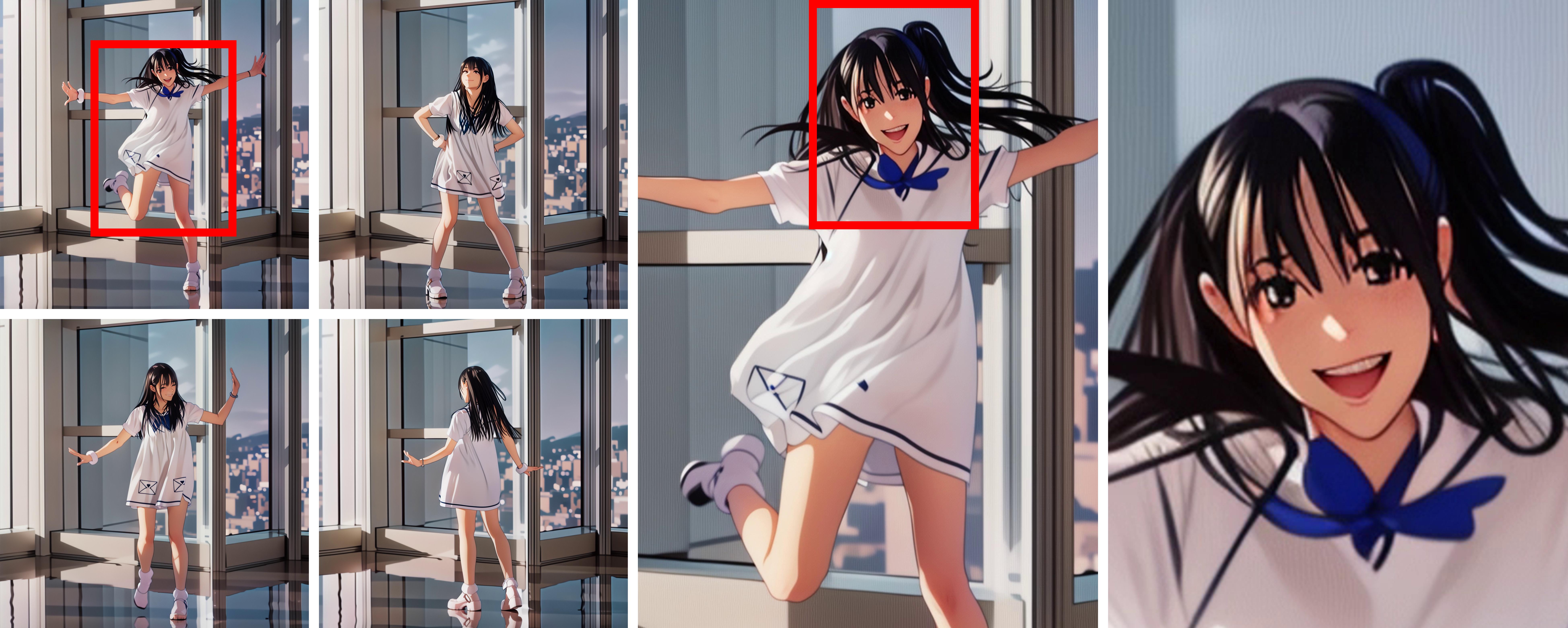} &
  \includegraphics[width=0.475\linewidth]{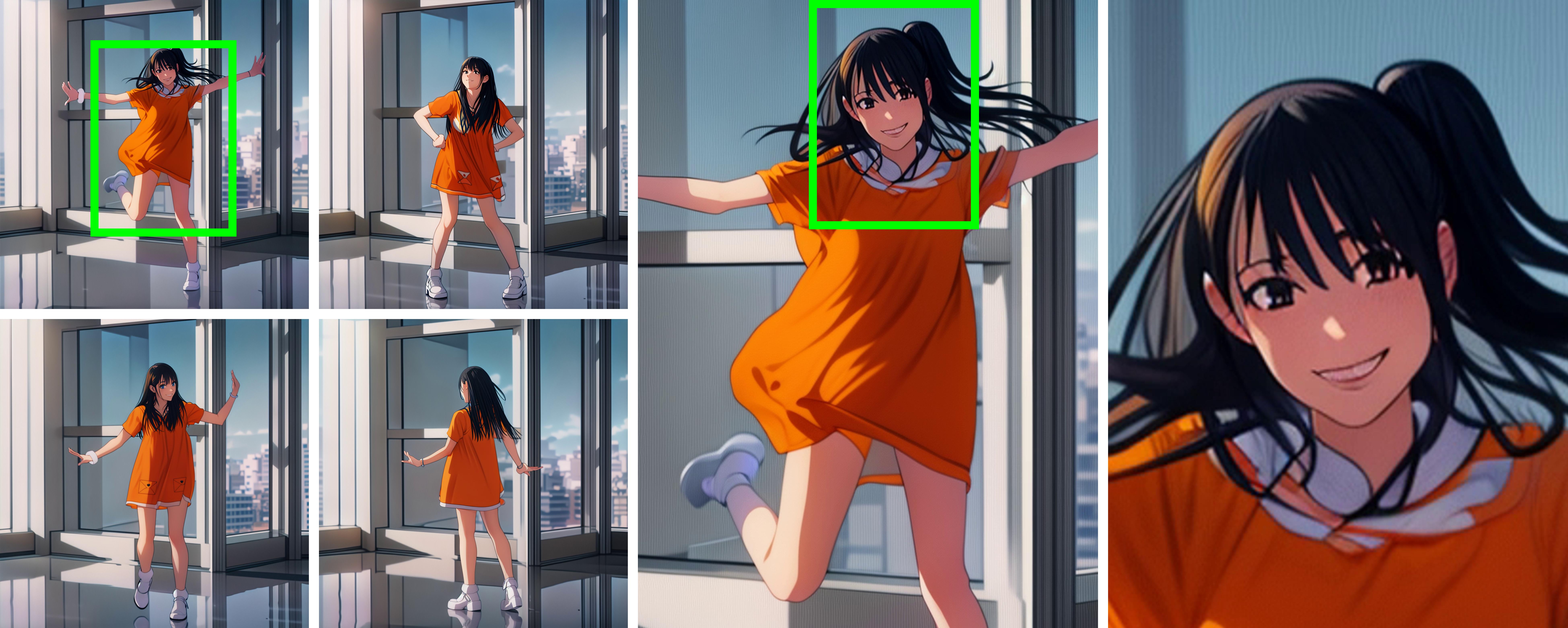} \\
  (g) Diffutoon (toon shading) & (h) Diffutoon (toon shading with editing signals) \\
  \end{tabular}
  \caption{Visual comparison with other methods. The prompt used for editing is ``best quality, perfect anime illustration, a girl is dancing, smile, solo,
  \colorbox[rgb]{0.9,0.9,0.9}{
      \colorbox[rgb]{0.98,0.51,0.12}{\textcolor[rgb]{1,1,1}{orange dress}},
      \colorbox[rgb]{0.10,0.10,0.10}{\textcolor[rgb]{1,1,1}{black hair}},
      \colorbox[rgb]{1.00,1.00,1.00}{\textcolor[rgb]{0,0,0}{white shoes}},
      \colorbox[rgb]{0.46,0.71,0.78}{\textcolor[rgb]{1,1,1}{blue sky}}
  }
  ''. Since the resolution of our generated video is extremely high, we enlarge some areas to view details. We highly recommend readers to see the videos on our project page.}
  \label{figure:case_study}
\end{figure*}

\begin{figure*}[htbp]
  \centering
  \begin{tabular}{cc}
  \includegraphics[width=0.475\linewidth]{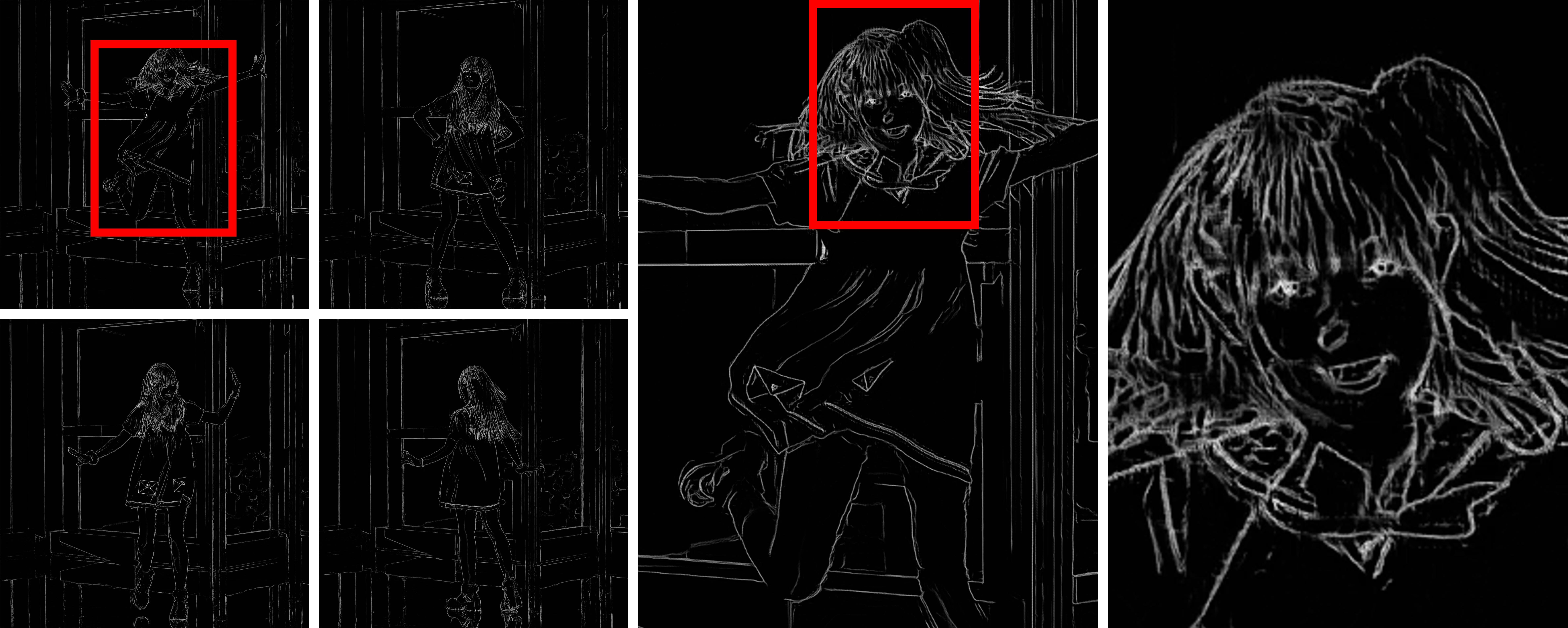} &
  \includegraphics[width=0.475\linewidth]{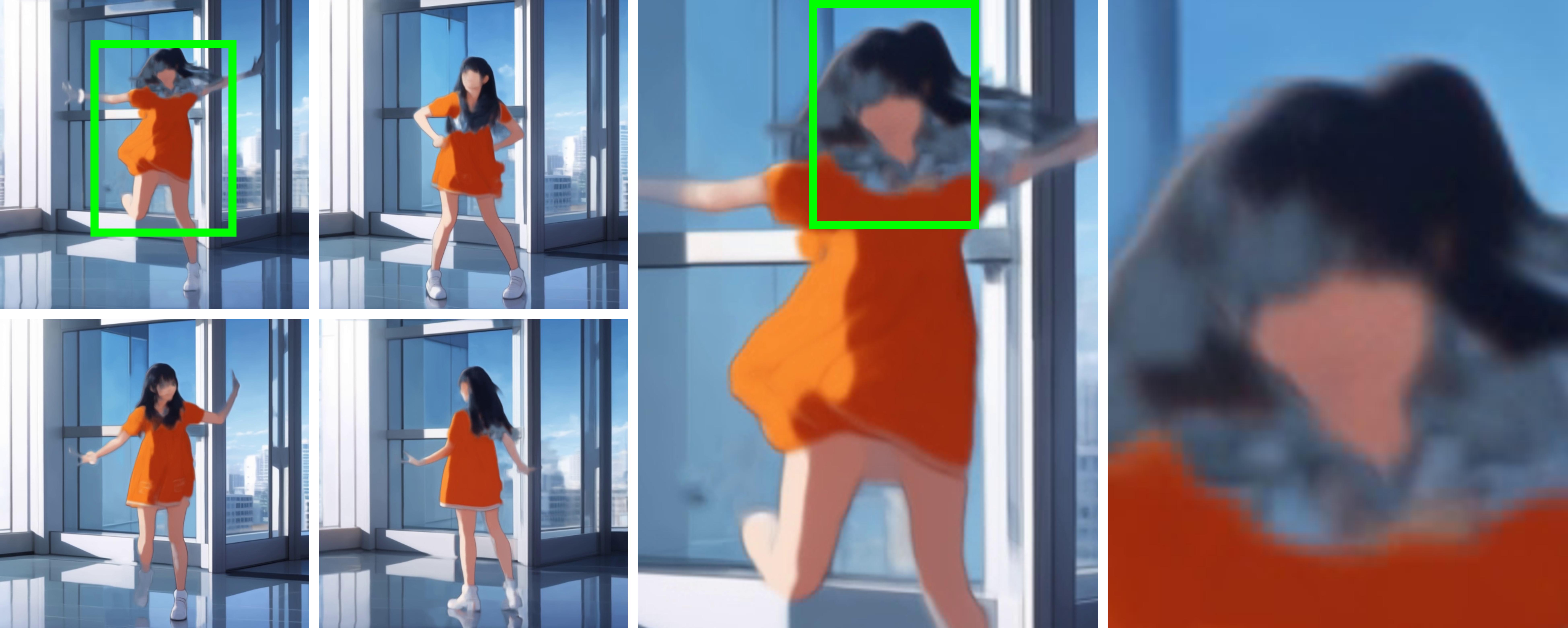} \\
  (a) Outline video & (b) Generated color video \\
  \end{tabular}
  \caption{Intermediate results of Diffutoon. In the main toon shading pipeline, the video is synthesized according to the outline video and the color video. When the editing branch is enabled, the generated color video contains the editing signals.}
  \label{figure:intermediate_results}
\end{figure*}

Currently, finding accurate evaluation metrics to measure video quality remains challenging, and there has been some controversy in recent years concerning evaluation metrics \cite{brooks2022generating,blattmann2023align,ouyang2023codef}. In our experiments, we evaluate the quality of videos generated by each method in three aspects. 1) \textbf{Aesthetics}: Visual appeal is quantified through an aesthetic score \cite{schuhmann2022laion}, providing a measure of the overall visual quality of the generated videos. 2) \textbf{Text-video similarity}: To evaluate the relevance of generated videos to the given text in the toon shading with editing signals task, we use the cosine similarity calculated by the CLIP model \cite{radford2021learning}. 3) \textbf{Video consistency}: Evaluating video consistency is challenging. While earlier studies \cite{wang2023zero,qi2023fatezero} commonly utilized feature similarity of adjacent frames, this approach is limited by embeddings computed by the CLIP model, which is specifically designed for images with a resolution of $224\times 224$. Therefore, this metric is not suitable for our experiments. Following Rerender-a-video \cite{yang2023rerender} and Pix2Video \cite{ceylan2023pix2video}, we adopt pixel-MSE as a metric for video consistency. Pixel-MSE is the mean square error between the warped frame and its corresponding target frame, where the warped frame is computed according to the estimated optical flow \cite{teed2020raft}. Note that the services provided by DomoAI and Gen-1 can only support 24 fps, which is not aligned with the original video. Consequently, the calculation of pixel-MSE for these two methods is not feasible. The quantitative results are shown in Table \ref{table:quantitative_results}. Our approach significantly surpasses other baseline models in both two tasks. The experimental results demonstrate the effectiveness of our method.

\begin{table}[t]
\centering
\tabcolsep=0.32em
\begin{tabular}{l|l|cc}
\hline
\multirow{2}{*}{Task}          & \multirow{2}{*}{Baseline} & \multicolumn{2}{c}{Preference} \\ \cline{3-4} 
                               &                           & Diffutoon          & Other     \\ \hline
\multirow{2}{*}{Toon shading}  & Rerender-a-video          & \textbf{98.21\%}   & 1.79\%    \\
                               & DomoAI                    & \textbf{90.77\%}   & 9.23\%    \\ \hline
\multirow{3}{*}{\begin{tabular}[c]{@{}l@{}}Toon shading\\ with editing signals\end{tabular}} & Rerender-a-video          & \textbf{97.44\%}   & 2.56\%    \\
                               & DomoAI                    & \textbf{82.35\%}   & 17.65\%   \\
                               & Gen-1                     & \textbf{95.74\%}   & 4.26\%    \\ \hline
\end{tabular}
\caption{User preference in human evaluation.}
\label{table:human_evaluation}
\end{table}

In addition to using the aforementioned metrics to evaluate each method, we also conducted a human evaluation involving 10 participants. In each evaluation episode, each participant is presented with two videos, one generated by our method and the other generated by a randomly selected baseline method. Participants are asked to choose the video with the better visual effects. We recorded the proportion of participant choices in Table \ref{table:human_evaluation}. Among these results, it is evident that users overwhelmingly believe that our method is capable of producing videos with superior visual effects. This further demonstrates the superiority of our approach.

\subsection{Case Study}

\begin{figure}[tbp]
  \centering
  \includegraphics[width=0.99\linewidth]{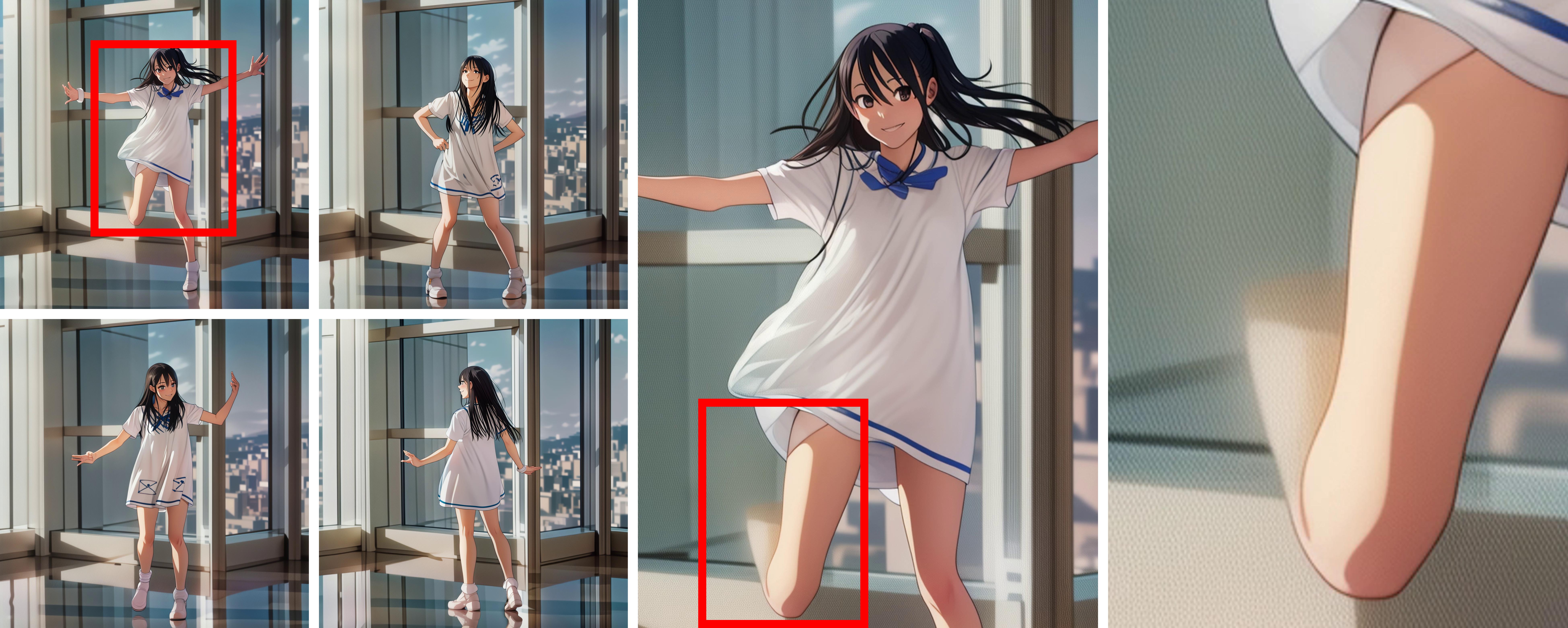}
  \caption{Video rendered without outline information.}
  \label{figure:without_outline}
\end{figure}

\begin{figure}[tbp]
  \centering
  \includegraphics[width=0.99\linewidth]{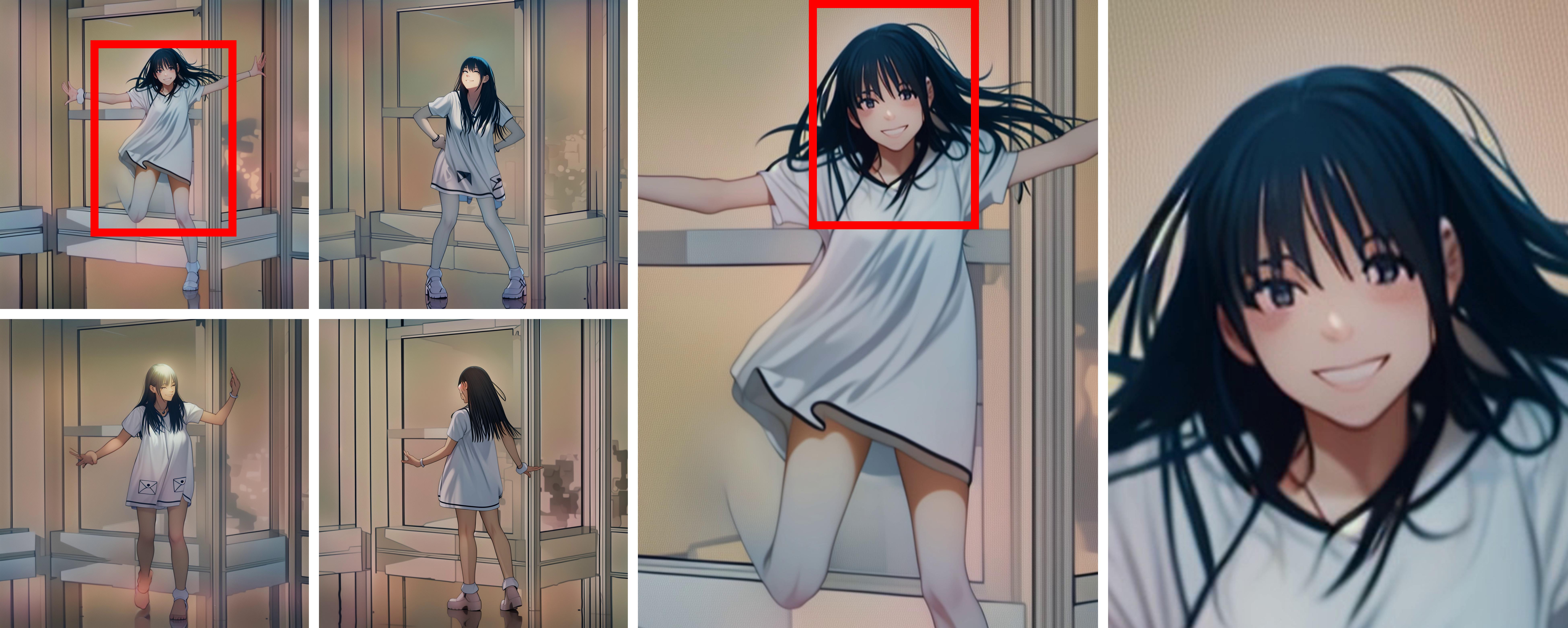}
  \caption{Video rendered without color information.}
  \label{figure:without_color}
\end{figure}

Figure \ref{figure:case_study} presents video samples generated by different methods. In the original video (Figure \ref{figure:case_study}a), a girl is dancing with fast movements, posing a significant challenge for each video processing method. Gen-1 and Rerender-a-video struggle to effectively handle high-resolution videos, resulting in facial distortions of the character. In the videos generated by DomoAI (Figure \ref{figure:case_study}e and Figure \ref{figure:case_study}f), there is missing content in the third frame, and the character's movements in the fourth frame do not align with the original video. This indicates that DomoAI cannot accurately capture motion features from the original video and reproduce the character's pose. Contrastingly, videos generated by Diffutoon (Figure \ref{figure:case_study}g and Figure \ref{figure:case_study}h) showcase the preservation of details such as lighting, hair, and pose, while maintaining a visual style closely aligned with anime aesthetics. Notably, in the toon shading with editing signals task, our method successfully achieves precise control based on the color information from the given text. These results intuitively highlight the robustness and efficacy of our approach.

In Figure \ref{figure:intermediate_results}, we present the intermediate results of Diffutoon, including the outline video and the color video generated by the editing branch. The outline video precisely retains the structural information for rendering an anime-style frame, ensuring the visual quality. The generated color video exhibits blurriness due to the rapid motion of the dancing girl. It implies that the editing branch, when operating independently, fails to produce a video of high quality. The outline video and the color video provide essential information for rendering a high-resolution video in Figure \ref{figure:case_study}h. For more video examples, please see the project page.

\subsection{Ablation Study}

Since the effectiveness of the motion modules has been widely evaluated by prior work \cite{xing2023survey}, we further investigate the effectiveness of the two ControlNet models in the main toon shading pipeline. The rendered videos without each ControlNet model are shown in Figure \ref{figure:without_outline} and Figure \ref{figure:without_color}. The lack of outline information results in mutilation within the frame. The lack of color guidance results in poor visual quality, with noticeable flickering on the head. It proves that the outline and color are both essential.

\begin{figure}[tbp]
  \centering
  \includegraphics[width=0.99\linewidth]{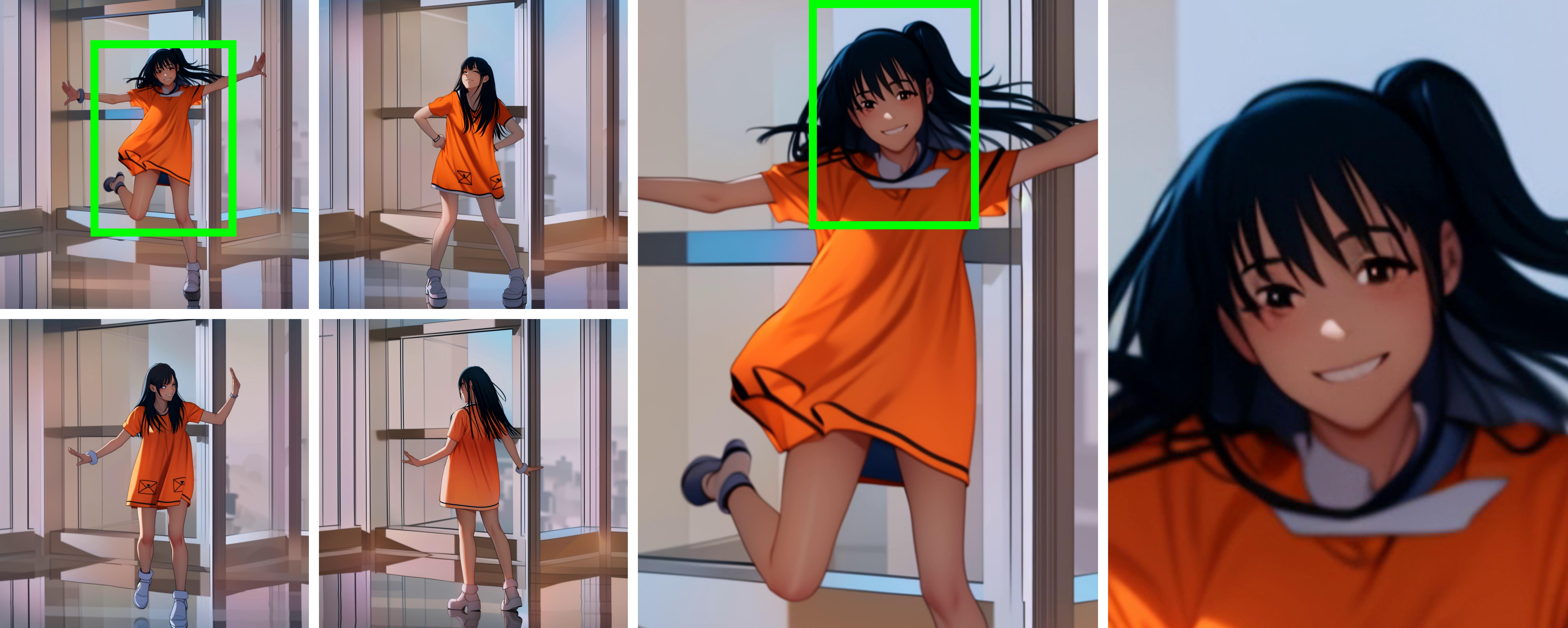}
  \caption{Video rendered by the editing branch with AnimateDiff.}
  \label{figure:single_pipeline}
\end{figure}

In the toon shading with editing signals task, we design an alternative approach that only contains a single pipeline. This approach is constructed based on the editing branch, wherein we replace FastBlend with AnimateDiff. The video generated by this approach is presented in Figure \ref{figure:single_pipeline}. We observe that this video is dark and lacks aesthetic appeal. As we mentioned in Section \ref{subsection:editing_branch}, the reason is that AnimateDiff relies on a modified DDIM scheduler. This scheduler is inconsistent with the Stable Diffusion backbone and the inconsistency is detrimental for synthesizing high-quality videos. However, this pitfall has minimal influence on the main toon shading pipeline, because the color is fixed by the ControlNet. It proves the necessity of maintaining a separate pipeline architecture.

\section{Conclusion and Future Work}

In this paper, we investigate an innovative form of toon shading based on diffusion models, intending to directly transmute photorealistic videos into anime styles. We introduce an advanced toon shading approach which consists of a main toon shading pipeline and an editing branch. Our approach is capable of processing high-resolution long videos, and can also edit the video via the editing branch. The comprehensive experimental results demonstrate the efficacy of our approach. However, Diffutoon is a toon shading method, not a general video stylization method, as it cannot handle other styles (e.g., realistic, oil painting, and ink painting). In the future, we will focus on exploring more applications within the domain of video processing.

\bibliographystyle{ACM-Reference-Format}
\bibliography{sample-base}

\appendix

\section{Model Components}

After experimental testing, we decided to utilize several open-source models obtained from the open-source community. These models are listed in Table \ref{table:model_list}. Benefiting from the abundant open-source models, we succeed in designing such a fantastic toon shading pipeline. 

\begin{table*}[tp]
\begin{tabular}{l|ccl}
\hline
Model type            & \begin{tabular}[c]{@{}c@{}}Main\\ toon\\ shading\\ pipeline\end{tabular} & \begin{tabular}[c]{@{}c@{}}Video\\ editing\\ branch\end{tabular} & URL                                                                   \\ \hline
Stable Diffusion      & \checkmark                 & \checkmark           & \url{https://civitai.com/models/34553/aingdiffusion}                        \\
ControlNet (Outline)  & \checkmark                 &                      & \url{https://huggingface.co/lllyasviel/control_v11p_sd15_lineart}           \\
ControlNet (Color)    & \checkmark                 &                      & \url{https://huggingface.co/lllyasviel/control_v11f1e_sd15_tile}            \\
ControlNet (Softedge) &                            & \checkmark           & \url{https://huggingface.co/lllyasviel/control_v11p_sd15_softedge}          \\
ControlNet (Depth)    &                            & \checkmark           & \url{https://huggingface.co/lllyasviel/control_v11f1p_sd15_depth}           \\
Motion modules        & \checkmark                 &                      & \url{https://huggingface.co/guoyww/animatediff/blob/main/mm_sd_v15_v2.ckpt} \\
Textual inversion     & \checkmark                 & \checkmark           & \url{https://civitai.com/models/11772}                                      \\ \hline
\end{tabular}
\caption{List of models utilized in Diffutoon.}
\label{table:model_list}
\end{table*}

\section{Parameter Settings}

\begin{table*}[tp]
\centering
\begin{tabular}{l|l|c}
\hline
Components & Parameter             & Value \\ \hline
\multirow{9}{*}{\begin{tabular}[c]{@{}l@{}}Main\\ toon\\ shading\\ pipeline\end{tabular}}
& frame height                   & 1536  \\
& frame width                    & 1536  \\
& classifier-free guidance scale & 7     \\
& denoising strength             & 1     \\
& inference steps                & 10    \\
& sliding window size            & 16    \\
& sliding window stride          & 8     \\
& conditioning scale (outline)   & 0.5   \\
& conditioning scale (color)     & 0.5   \\ \hline
\multirow{9}{*}{\begin{tabular}[c]{@{}l@{}}Video\\ editing\\ branch\end{tabular}}
& frame height                   & 512   \\
& frame width                    & 512   \\
& classifier-free guidance scale & 7     \\
& denoising strength             & 0.9   \\
& inference steps                & 20    \\
& sliding window size            & 8     \\
& sliding window stride          & 4     \\
& conditioning scale (depth)     & 0.5   \\
& conditioning scale (softedge)  & 0.5   \\ \hline
\multirow{7}{*}{FastBlend}
& inference mode                 & accurate \\
& sliding window size            & 30       \\
& batch size                     & 64       \\
& tracking mechanism             & enabled  \\
& patch size                     & 5        \\
& number of iterations           & 5        \\
& guide weight $\alpha$          & 10       \\ \hline
\end{tabular}
\caption{Parameter settings in the experiments.}
\label{table:parameter}
\end{table*}

The parameter settings of our approach are detailed in Table \ref{table:parameter}. Since our approach has a robust tolerance to color video, we use a lower resolution and sliding window size in the editing branch for faster generation. The denoising strength quantifies the extent of noise introduced into the video, with a value of 1 indicating complete frame replacement and rerendering, while 0 implies no modifications to the video. In the editing branch, we set the denoising strength to 0.9, retaining a little information from the input video. The number of inference steps is 20 in the editing branch, which is larger than that of the main toon shading pipeline. This adjustment is based on empirical findings that fewer steps may lead to frames that are misaligned with the desired editing prompt. These parameters are manually tuned to optimize speed without compromising the resulting quality. For the parameters associated with FastBlend, the accurate inference mode is utilized, and readers can refer to the original paper \cite{duan2023fastblend} for more comprehensive details on these configurations.

\end{document}